\documentclass[lettersize,journal]{IEEEtran}
\usepackage{amsmath,amsfonts}
\usepackage{algorithmic}
\usepackage{algorithm}
\usepackage{array}
\usepackage[caption=false,font=normalsize,labelfont=sf,textfont=sf]{subfig}
\usepackage{textcomp}
\usepackage{stfloats}
\usepackage{url}
\usepackage{verbatim}
\usepackage{graphicx}
\usepackage{cite}

\usepackage{xcolor}
\usepackage{multirow}
\usepackage{graphicx}
\usepackage{booktabs}
\usepackage{tabularx}
\usepackage[flushleft]{threeparttable}
\usepackage{array}
\usepackage{pifont} 
\usepackage{booktabs}
\usepackage{textcomp}
\usepackage{makecell}
\usepackage{caption}
\usepackage{hyperref}
\usepackage{bm}
\usepackage[table]{xcolor}
\definecolor{mygreen}{HTML}{77a461}
\definecolor{myblue}{HTML}{6c8ebf}
\definecolor{myred}{HTML}{b85450}

\begin{document}
\bstctlcite{IEEEexample:BSTcontrol}

\title{DiffVC-ONE: Diffusion-based Generative Video Compression\\with One-Step Video Diffusion Transformer}

\author{Wenzhuo Ma, Zhenzhong Chen
\thanks{Wenzhuo Ma and Zhenzhong Chen are with the school of Remote Sensing and Information Engineering, Wuhan University, Hubei 430079, China. Corresponding author: Zhenzhong Chen, E-mail:zzchen@ieee.org}}

\maketitle

\begin{abstract}
  Generative video compression can recover rich visual details at low bitrates, but simultaneously achieving high temporal consistency and low inference cost remains challenging. To address this issue, we propose DiffVC-ONE, a diffusion-based generative video compression framework built on a one-step Video Diffusion Transformer. First, we introduce a Unified Unidirectional Latent Compressor that uses a shared model to efficiently and uniformly compress compact latent slices. We then develop a Video DiT-based One-Step Diffusion Enhancer that uses the reconstructed latent slices as content anchors and performs single-step spatio-temporal perceptual enhancement over an entire group of pictures. Finally, a Hybrid Condition Generator extracts structural, strength, and semantic conditions from the reconstructed content and quantization information. These conditions preserve faithful regions, control the degree of generative enhancement, and supplement content-aware perceptual details during one-step diffusion enhancement. Extensive experiments on multiple standard benchmarks demonstrate that DiffVC-ONE achieves state-of-the-art perceptual quality and temporal consistency with low inference cost.
\end{abstract}

\begin{IEEEkeywords}
  Generative Video Compression, Video Diffusion Transformer, One-Step Diffusion.
\end{IEEEkeywords}

\section{Introduction}
\IEEEPARstart{V}{ideo} has become a dominant form of Internet traffic and plays a crucial role in applications such as online conferencing and immersive media, making efficient video coding increasingly important. Traditional standards, including H.264/AVC~\cite{AVC}, H.265/HEVC~\cite{HEVC}, and H.266/VVC~\cite{VVC}, have achieved remarkable compression performance through decades of development. However, their hand-crafted modules are difficult to optimize jointly for the overall rate-distortion objective, leaving limited room for further improvement. In recent years, Neural Video Compression (NVC) offers a new paradigm by learning spatial and temporal redundancies directly from data. With learnable motion estimation, context modeling, and entropy coding, recent NVC methods~\cite{LSTVC,DCVC-DC,DCVC-FM,DCVC-RT,ECVC,DCMVC,SEVC,UI2C,DCVC-UF,CAHH-NVC,UniLVC,DCVC-B,BRHVC,BiECVC,HR-NVC} have surpassed traditional codecs on several standard datasets~\cite{HEVC,MCL-JCV,UVG,USTC-TD}. Nevertheless, most existing methods are primarily optimized using MSE or MS-SSIM~\cite{MS-SSIM}. Such distortion-oriented objectives tend to produce averaged reconstructions, resulting in over-smoothed textures, blurred edges, and missing high-frequency details, particularly at low bitrates. Consequently, their subjective quality remains unsatisfactory.

\begin{figure}[!t]
  \centering
  \includegraphics[width=\columnwidth]{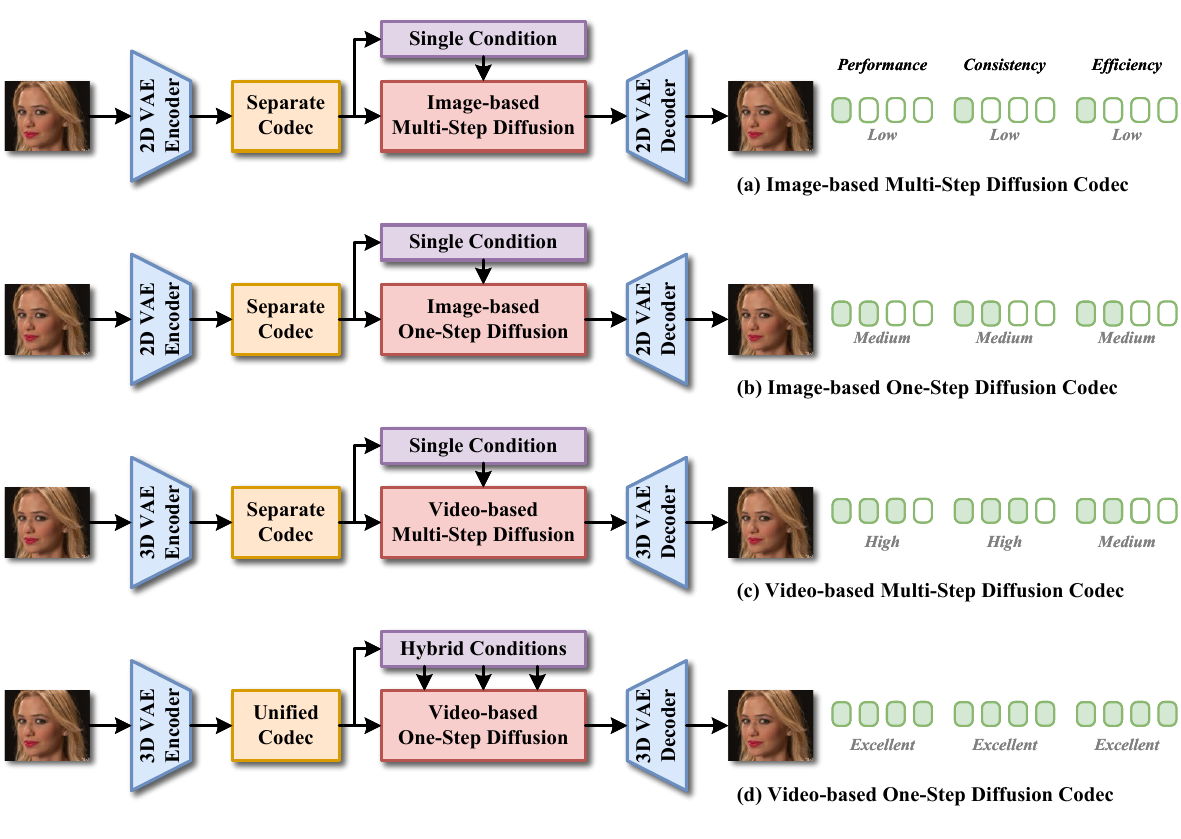}
  \caption{
    Overview of diffusion-based video compression paradigms. Existing methods include image-based multi-step, image-based one-step, and video-based multi-step codecs. In contrast, DiffVC-ONE belongs to the video-based one-step paradigm, achieving a favorable trade-off among compression performance, temporal consistency, and inference efficiency.
  }
  \label{fig:overview}
\end{figure}

To better balance bitrate, distortion, and perceptual quality, perceptual NVC~\cite{DVC-P,PLVC,GLC-video,GLVC,PNVC} has received increasing attention. Meanwhile, pre-trained diffusion models~\cite{SD,DiT} have demonstrated powerful generative priors in low-level vision, offering a promising direction for perceptual video compression. By introducing diffusion models at the decoder, this paradigm restores texture details lost during compression and produces perceptually richer reconstructions. Recent studies~\cite{DiffVC,DiffVC-OSD,S2VC,GNVC-VD,YODA} have explored diffusion sampling, condition modeling, and video generative priors. However, existing methods still face several limitations:

\paragraph{Separate coding architectures} Existing methods typically use separate codecs for intra and inter frames. DiffVC~\cite{DiffVC}, DiffVC-OSD~\cite{DiffVC-OSD}, S2VC~\cite{S2VC} and YODA~\cite{YODA}, for example, employ an independent image codec for intra frames and another codec for inter frames. Differences in distortion distributions and perceptual characteristics between the two codecs may cause temporal discontinuities across frame types. Although GNVC-VD~\cite{GNVC-VD} performs compression in the latent space, it still requires an additional ELIC-based image codec for intra latents, increasing model parameter redundancy. Recent methods such as UI2C~\cite{UI2C} and UniLVC~\cite{UniLVC} have demonstrated the feasibility of unified pixel-domain codecs. However, a unified and efficient latent-domain codec for diffusion-based video compression remains largely unexplored.

\paragraph{Trade-off between temporal consistency and decoding efficiency} Most existing methods~\cite{DiffVC,DiffVC-OSD,S2VC,YODA} rely on image diffusion models whose generative priors are limited to individual frames. Even with motion or cross-frame conditions, frame-wise reconstruction may introduce stochastic texture variations, causing flickering, detail drift, and semantic inconsistency over time. GNVC-VD~\cite{GNVC-VD} instead employs a video diffusion model to jointly reconstruct multiple frames, offering better temporal consistency. However, its multi-step iterative sampling incurs substantial computational cost and decoding latency. Therefore, efficiently exploiting the temporally coherent priors of video diffusion models without iterative sampling remains a key challenge.

\paragraph{Insufficient conditional guidance} In diffusion-based video compression, the diffusion model acts as a condition-driven generative enhancer rather than generating from random noise. It uses compressed reconstructions as content anchors to recover perceptual details lost under bitrate constraints, making reconstruction quality highly dependent on the sufficiency of the conditions. Existing methods typically rely on reconstructed latents, fixed text prompts, or a single visual feature, which cannot jointly represent spatial structure, bitrate-dependent enhancement strength, and content semantics. Such insufficient guidance enlarges the solution space and increases the risk of hallucinated or implausible details.

To address the above issues, we propose \textbf{DiffVC-ONE}, a diffusion-based generative video compression framework built upon a one-step Video Diffusion Transformer. As shown in Fig.~\ref{fig:overview}, unlike previous methods, DiffVC-ONE performs unified temporal compression, one-step video generative enhancement, and hybrid condition modeling in the compact latent space. Our main contributions are summarized as follows:

\begin{itemize}
    \item \textbf{Unified Unidirectional Latent Compressor (U2LC):} We introduce a shared coding architecture for all latent slices within a GOP, avoiding quality inconsistency, parameter redundancy, and training overhead caused by a separate intra-frame codec. Since the video VAE produces compact representations in both spatial and temporal dimensions, U2LC can effectively exploit their spatial-temporal correlations. We further investigate three representative reference structures: Unidirectional, Joint, and Bidirectional. Experiments show that the unidirectional variant achieves the best overall trade-off among compression performance, complexity, and flexibility, and is therefore adopted as the default mode in DiffVC-ONE.

    \item \textbf{Video DiT-based One-Step Diffusion Enhancer (OSDiT):} We leverage the temporally coherent generative prior of a pre-trained Video DiT and formulate video reconstruction as one-step conditional latent enhancement. Unlike frame-wise enhancement with image diffusion models, OSDiT processes an entire GOP and jointly restores spatiotemporal textures, improving cross-frame consistency. Since the reconstructed latent representation already preserves the main video structure, the diffusion model only needs to recover information lost during compression rather than generate from pure noise. This enables us to replace conventional multi-step sampling with a single diffusion step, substantially reducing the computational cost of generative enhancement.

    \item \textbf{Hybrid Condition Generator (HCG):} To provide sufficient guidance, we extract three complementary conditions--structural, strength, and semantic--from the reconstructed latents and quantization parameters. The structural condition describes the spatial layout, object boundaries, and temporal motion of the reconstructed video, serving as a reliable fidelity anchor. The strength condition dynamically controls the degree of generative enhancement according to the quantization level and latent distortion, making enhancement more conservative at high bitrates and more sufficient at low bitrates. The semantic condition extracts content-aware high-level representations, guiding the model to recover perceptual details that match the current video content. Together, these conditions reduce generative uncertainty and suppress implausible hallucinations during one-step diffusion.
\end{itemize}

Extensive experiments across multiple benchmarks demonstrate that DiffVC-ONE achieves state-of-the-art perceptual quality and temporal consistency, producing reconstructions with richer details and more pleasing visual quality. The remainder of this paper is organized as follows. Section~\ref{sec:related_work} reviews the related work. Section~\ref{sec:method} presents the proposed method in detail. Section~\ref{sec:experiments} reports the experimental results and analysis. Section~\ref{sec:conclusion} concludes the paper.

\section{Related Work}\label{sec:related_work}
\subsection{Neural Video Compression}
Neural Video Compression (NVC) leverages deep neural networks and end-to-end optimization to learn compact video representations. Early work DVC~\cite{DVC} replaced conventional coding modules with learnable networks, while DCVC~\cite{DCVC} introduced conditional coding that exploits temporal contexts from decoded frames for reconstruction and entropy modeling. Subsequent DCVC series~\cite{DCVC-TCM,DCVC-HEM,DCVC-DC,DCVC-FM,DCVC-RT} further advanced context and motion modeling, variable-rate coding, and computational efficiency. The latest DCVC-UF~\cite{DCVC-UF} further improves the inference efficiency while maintaining strong compression performance, through chunk-based processing and streamlined entropy coding. Beyond unidirectional reference structure, several methods investigate bidirectional architectures~\cite{IBVC,B-CANF,DCVC-B,BRHVC,BiECVC,HR-NVC} by exploiting bidirectional temporal contexts. For example, BRHVC~\cite{BRHVC} performs bidirectional motion and contextual fusion, while BiECVC~\cite{BiECVC} combines local and non-local contexts to outperform VTM-RA~\cite{VVC}. In addition, recent works have started to investigate unified frameworks, which no longer distinguish intra frames from inter frames~\cite{UI2C}, or even support All-Intra, Low-Delay, and Random-Access coding modes with a single framework~\cite{UniLVC}. However, the above NVC methods are mainly optimized for rate-distortion objectives. Under the rate-distortion-perception trade-off~\cite{RDP}, they tend to produce over-smoothed and blurred reconstructions with limited perceptual quality.

\begin{figure*}[!t]
  \centering
  \includegraphics[width=2\columnwidth]{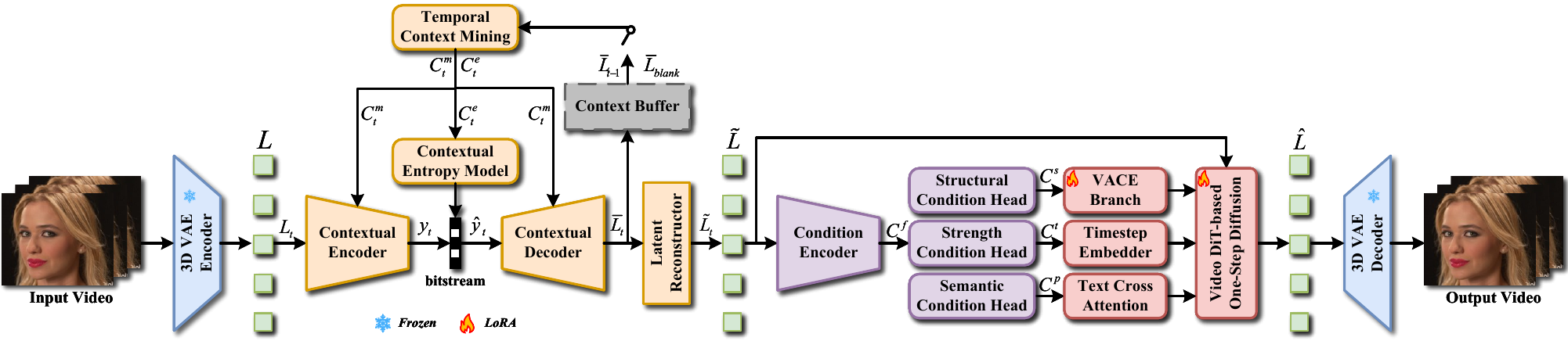}
  \caption{
    The framework of the proposed DiffVC-ONE.
  }
  \label{fig:framework}
\end{figure*}

\subsection{Generative Video Compression}
Generative video compression can be broadly categorized into GAN-based and diffusion-based approaches. GAN-based NVC~\cite{PLVC,GLC-video,GLVC,PNVC} employs adversarial training~\cite{GAN} to improve texture richness and visual realism, but its training can be unstable and may introduce unnatural artifacts. Diffusion-based NVC~\cite{DiffVC,I2VC,DiffVC-OSD,S2VC,GNVC-VD,YODA} typically incorporates pre-trained diffusion models into the decoder to recover details lost during compression. DiffVC~\cite{DiffVC} uses multi-step image diffusion with temporal information reuse, but frame-wise enhancement remains computationally expensive and prone to cross-frame texture inconsistency. DiffVC-OSD~\cite{DiffVC-OSD} improves efficiency through one-step frame-wise diffusion, yet cannot fully model spatio-temporal correlations. GNVC-VD~\cite{GNVC-VD} jointly enhances an entire GOP using a video diffusion model and achieves better temporal consistency, but still relies on costly multi-step sampling. Other diffusion-based frameworks, including LoRA compression~\cite{IVR-DFM} and diffusion trajectory compression~\cite{Free-GVC}, face similar efficiency limitations caused by multi-step sampling. Motivated by these observations, we propose a video compression framework based on a one-step Video Diffusion Transformer to better balance compression performance, temporal consistency, and inference efficiency.

\section{Method}\label{sec:method}
In this work, we propose DiffVC-ONE, whose overall framework is illustrated in Fig.~\ref{fig:framework}. It consists of four principal components: a pre-trained 3D VAE (blue), a Unified Unidirectional Latent Compressor (U2LC, orange), a Video DiT-based One-Step Diffusion Enhancer (OSDiT, red), and a Hybrid Condition Generator (HCG, purple). Given a group of pictures $X \in \mathbb{R}^{T\times H\times W\times 3}$, the 3D VAE Encoder produces compact latent slices $L \in \mathbb{R}^{(1+\frac{T-1}{4})\times\frac{H}{8}\times\frac{W}{8}\times16}$. U2LC compresses $L$ into a bitstream and reconstructs $\tilde{L}$. HCG then derives complementary structural $C^s$, strength $C^t$, and semantic $C^p$ conditions from $\tilde{L}$ and related information. Guided by these conditions, OSDiT performs one-step diffusion enhancement to obtain perceptually improved and temporally consistent latent slices $\hat{L}$, which are finally decoded by the 3D VAE Decoder into the reconstructed video $\hat{X}$.

\subsection{Unified Unidirectional Latent Compressor}
\begin{figure*}[!t]
  \centering
  \includegraphics[width=1.5\columnwidth]{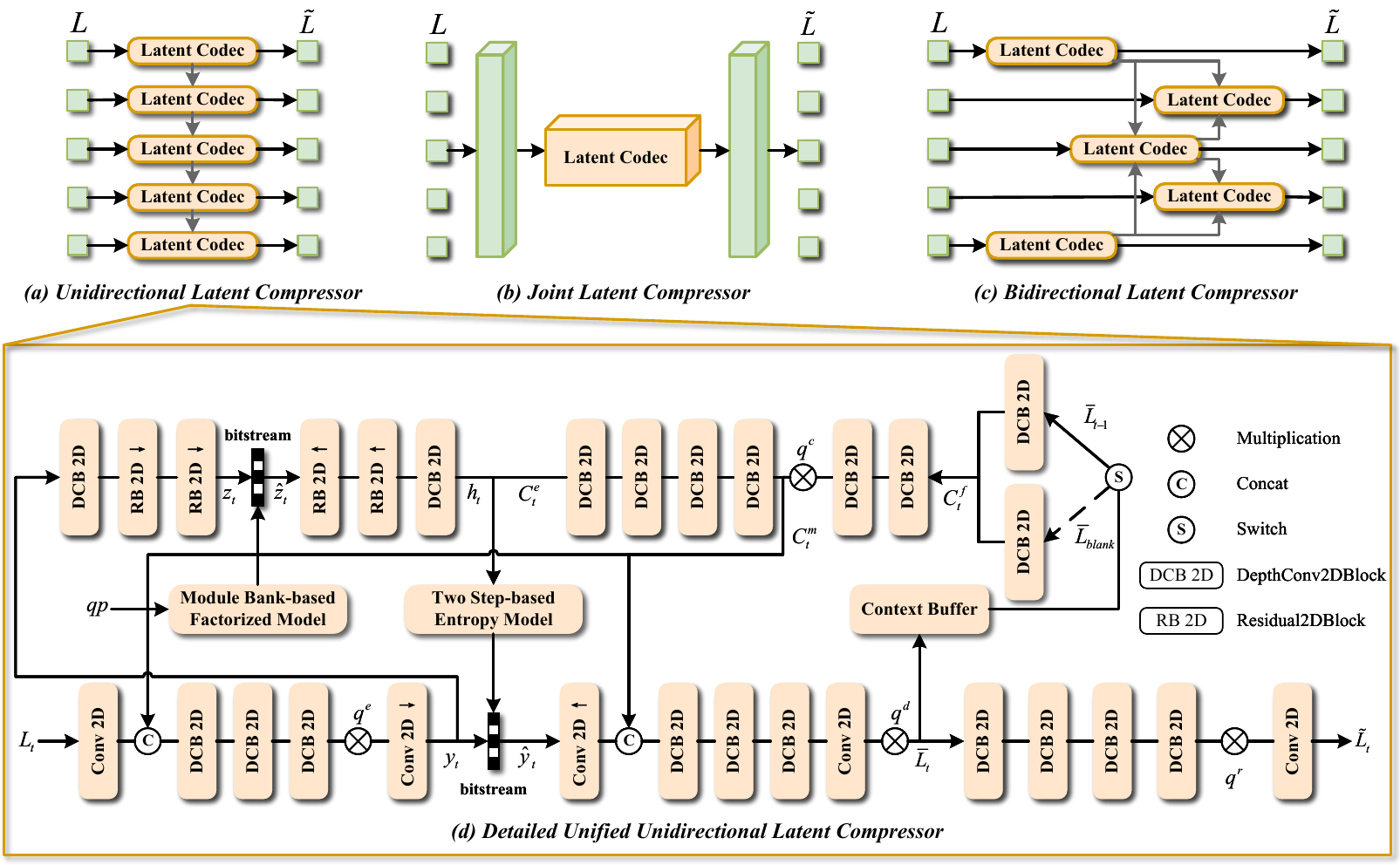}
  \caption{
    Latent Compressors under different reference structures and the detailed framework of the U2LC.
  }
  \label{fig:u2lc}
\end{figure*}
Existing diffusion-based video compression methods typically adopt either fully separate codecs, where an additional image codec handles intra frames, or partially separate codecs, where the intra latent slice is encoded independently. The former may produce heterogeneous reconstruction characteristics and quality fluctuations, while the latter increases model redundancy and training complexity. Inspired by unified distortion-oriented codecs~\cite{UI2C,UniLVC}, we propose the Unified Unidirectional Latent Compressor (U2LC), as shown in Fig.~\ref{fig:u2lc}.

Unlike pixel-domain codecs, U2LC operates on compact, low-resolution latent representations, reducing the need for explicit motion modeling and aligning with the implicit motion-modeling paradigm of DCVC-RT~\cite{DCVC-RT}. Accordingly, we adapt DCVC-RT to the latent space and unify intra- and inter-slice coding within a shared unidirectional architecture. For the $t$-th latent slice $L_t$, the Contextual Encoder extracts a compact representation conditioned on the temporal context $C_t^m$, followed by quantization and entropy coding. At the decoder, $\hat{y}_t$ is decoded into the contextual latent $\bar{L}_t$, which is cached for subsequent slices and refined by the Latent Reconstructor to obtain $\tilde{L}_t$:
\begin{equation}
  \begin{aligned}
    \hat{y}_t &= \operatorname{Quant}\left(\operatorname{ContextualEncoder}(L_t \mid C_t^m,q^e)\right), \\
    \bar{L}_t &= \operatorname{ContextualDecoder}(\hat{y}_t \mid C^m_t, q^d), \\
    \tilde{L}_t &= \operatorname{LatentReconstructor}(\bar{L}_t, q^r).
  \end{aligned}
\end{equation}
To estimate accurate probability distributions for entropy coding, U2LC jointly exploits global, temporal, and spatial priors. For the global prior, the Hyper Encoder maps $y_t$ to $z_t$, which is quantized and compressed by the Module Bank-based Factorized Model~\cite{DCVC-RT}. The reconstructed hyper-latent $\hat{z}_t$ is then decoded into the global prior $h_t$:
\begin{equation}
  \begin{aligned}
    \hat{z}_t &= \operatorname{Quant}(\operatorname{HyperEncoder}(y_t)), \\
    h_t &= \operatorname{HyperDecoder}(\hat{z}_t).
  \end{aligned}
\end{equation}
For temporal modeling, the intra slice $L_0$ uses a blank latent $\bar{L}_{\mathrm{blank}}$, while each inter slice uses the preceding reconstruction $\bar{L}_{t-1}$ as temporal context. Separate lightweight adaptors map them to a unified feature $C_t^f$, enabling shared processing of intra and inter slices. Temporal Context Mining then generates $C_t^m$ for the main transform and $C_t^e$ for entropy modeling. Finally, the two-step entropy model~\cite{DCVC-RT} combines the temporal prior $C_t^e$ with the global prior $h_t$ and applies checkerboard spatial-context modeling to predict the Gaussian parameters $(\mu_t,\sigma_t)$:
\begin{equation}
  \begin{aligned}
    C^f_t &=
    \begin{cases}
      \operatorname{IntraAdaptor}(\bar{L}_{\mathrm{blank}}), & t = 0, \\
      \operatorname{InterAdaptor}(\bar{L}_{t-1}), & t > 0,
    \end{cases} \\
    (C^m_t, C^e_t) &= \operatorname{TemporalContextMining}(C^f_t, q^c), \\
    (\mu_t, \sigma_t) &= \operatorname{TwoStepEntropyModel}(h_t, C^e_t).
  \end{aligned}
\end{equation}
Here, $q^e$, $q^d$, $q^r$, and $q^c$ are channel-wise scaling parameters following DCVC-FM~\cite{DCVC-FM}, enabling variable-rate coding.

For reference structure, we implement Unidirectional, Joint, and Bidirectional variants within the Latent Compressor, as shown in Fig.~\ref{fig:u2lc}. The joint variant jointly models an entire GOP using a computationally expensive 3D architecture, while encoding multiple frames with a single model may create a representational bottleneck. The bidirectional variant employs hierarchical bidirectional references and, although it offers greater compression potential in principle, struggles to handle large-motion cross-layer references under a limited-complexity implicit motion-modeling framework. In contrast, the unidirectional variant recursively references previously decoded latent slices and naturally supports variable-length inputs. Considering compression performance, complexity, and flexibility, we adopt unidirectional variant as the default mode of DiffVC-ONE. Further analysis is provided in Sec.~\ref{sec:ablation_u2lc}.

\subsection{Video DiT-based One-Step Diffusion Enhancer}
The choice of generative prior is critical to diffusion-based NVC, as it largely determines the reconstruction capability of the overall framework. As illustrated in Fig.~\ref{fig:overview}, existing methods mainly employ image-based multi-step, image-based one-step, or video-based multi-step diffusion models. Image-based models can enhance individual-frame details but often struggle to maintain temporal consistency, whereas video diffusion models jointly capture spatiotemporal dependencies and provide more coherent reconstructions. However, conventional multi-step sampling incurs substantial inference latency. Since the latent slices reconstructed by the compressor already preserve the main video content and structure, iterative refinement from noise is unnecessary. We therefore adopt a video-based one-step diffusion prior and develop the Video DiT-based One-Step Diffusion Enhancer (OSDiT) for perceptually high-quality and temporally consistent reconstruction.

Specifically, the reconstructed latent slices $\tilde{L}$ are jointly enhanced by OSDiT in a single step to obtain $\hat{L}$:
\begin{equation}
  \hat{L} = \tilde{L} - \sigma\, v_{\theta}(\tilde{L}; C^s, \tau, C^p).
\end{equation}
Here, $v_{\theta}$ denotes the pre-trained Video DiT, while $C^s$ and $C^p$ denote the structural and semantic conditions generated by the Hybrid Condition Generator, respectively. The sampling coefficient $\sigma$ and timestep $\tau$ are determined by strength condition $C^t$. Further details are provided in Sec.~\ref{sec:hcg}.

\subsection{Hybrid Condition Generator} \label{sec:hcg}
\begin{figure}[!t]
  \centering
  \includegraphics[width=\columnwidth]{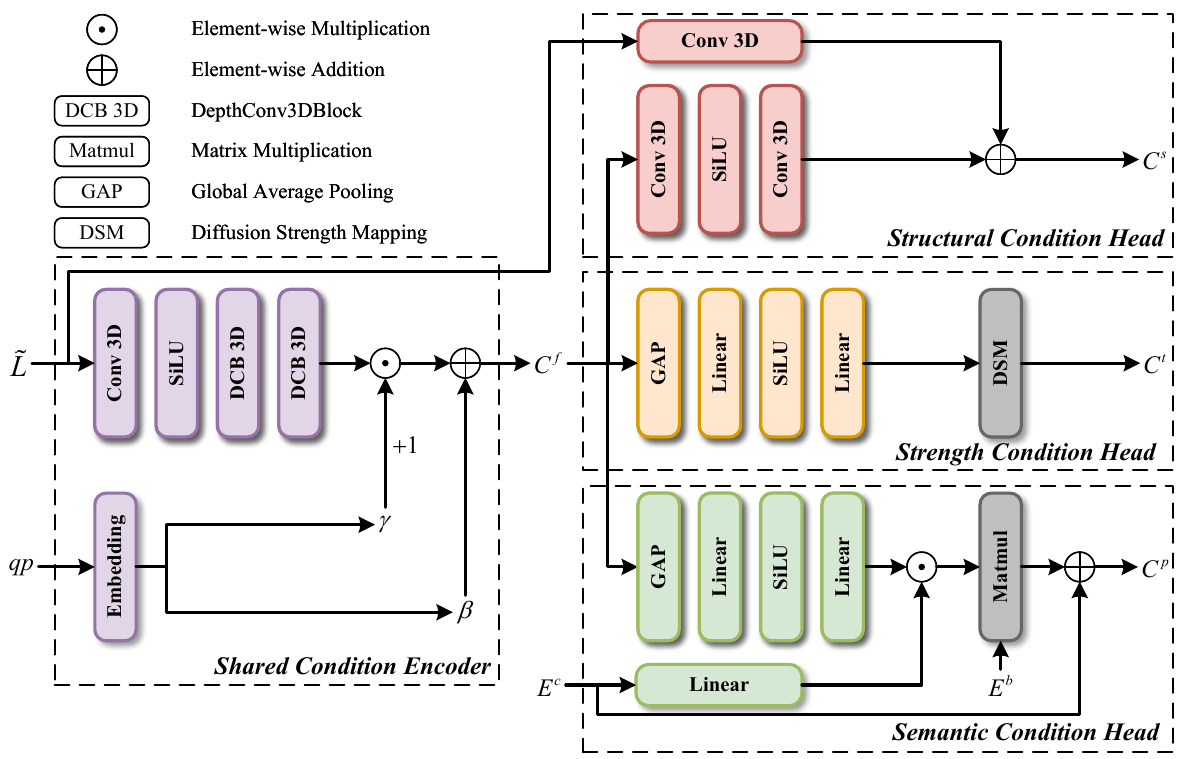}
  \caption{
    The architecture of Hybrid Condition Generator.
  }
  \label{fig:hcg}
\end{figure}

Diffusion-based video compression is fundamentally a condition-guided enhancement process. However, most existing methods rely on a single condition, which may provide insufficient guidance across varying bitrates and video content. To address this limitation, we propose the Hybrid Condition Generator (HCG), as illustrated in Fig.~\ref{fig:hcg}, which generates complementary structural, strength, and semantic conditions for one-step diffusion enhancement.

Given the reconstructed latents $\tilde{L}$, HCG first extracts shared spatiotemporal features $C^f$ using the lightweight 3D convolutional blocks $\mathcal{B}$. To make the conditions aware of the bitrate constraint, the quantization parameter $qp$ is embedded by $\mathcal{E}_q$ into channel-wise scale and bias for affine modulation: 
\begin{equation}
  (\gamma, \beta) = \mathcal{E}_q(qp), \qquad
  C^f = (1 + \gamma) \odot \mathcal{B}(\tilde{L}) + \beta,
\end{equation}

For the \textbf{structural condition} $C^s$, we add a linear projection of $\tilde{L}$ obtained by a single convolution $\mathcal{P}_s$ to a residual predicted from $C^f$ by a multi-layer transform network $\mathcal{D}_s$:
\begin{equation}
  C^s = \mathcal{P}_s(\tilde{L}) + \mathcal{D}_s(C^f).
\end{equation}
This residual design preserves reliable spatial and temporal structures while correcting local compression distortions. The resulting $C^s$ serves as the visual condition of OSDiT, providing a fidelity constraint for one-step enhancement.

For the \textbf{semantic condition} $C^p$, HCG first applies global average pooling to $C^f$ and predicts a rank-$r$ semantic code $a$. Let $E^c\in\mathbb{R}^{N\times d}$ denote the embedding of a constant prompt, where $N$ is the number of tokens and $d$ is the token dimension. With the text projection matrix $W_p\in\mathbb{R}^{d\times r}$ and a learnable semantic basis $E^b\in\mathbb{R}^{r\times d}$, HCG obtains a content-adaptive semantic condition through low-rank residual modulation:
\begin{equation}
  \begin{aligned}
    a &= \mathcal{H}_p(\operatorname{Pool}(C^f)), \\
    C^p  &= E^c + \bigl[(E^c W_p) \odot (\mathbf{1}_N a^\top)\bigr] E^b,
  \end{aligned}
\end{equation}
This design retains the prior of the constant prompt while efficiently injecting content-aware semantics, guiding OSDiT to recover perceptual details consistent with the current scene.

Finally, the \textbf{strength condition} $C^t$ is predicted from the pooled feature and mapped to the sampling strength $\sigma$ and corresponding timestep $\tau$:
\begin{equation}
  \begin{aligned}
    C^t &= \mathcal{H}_t(\operatorname{Pool}(C^f)), \\
    \sigma &= \sigma_{base} + (\sigma_{\max}-\sigma_{\min})\tanh(C^t), \\
    \tau &= \sigma N_{\mathrm{train}},
  \end{aligned}
\end{equation}
Here, $\sigma_{\mathrm{base}}$ is the predefined base strength, $[\sigma_{\min},\sigma_{\max}]$ denotes the valid range, and $N_{\mathrm{train}}$ is the number of diffusion training timesteps. By controlling $\sigma$ and $\tau$, HCG adaptively adjusts the contribution of the OSDiT prior, applying conservative enhancement when the reconstruction is reliable and stronger compensation when more information is lost. Overall, $C^s$, $C^p$, and $C^t$ constrain structural fidelity, semantic consistency, and enhancement strength, respectively, thereby connecting U2LC with OSDiT for effective reconstruction.

\subsection{Training Strategy}
To train model efficiently, we adopt a three-stage strategy.
In the first stage, U2LC is trained in the latent space to establish basic compression capability using a rate-distortion objective:
\begin{equation}
  Loss_1 = R + \lambda MSE(\tilde{L},L)
\end{equation}
where $R$ is the rate cost, $L$ and $\tilde{L}$ denote the original and reconstructed latent slices, respectively, and $\lambda$ balances rate and distortion.

In the second stage, OSDiT and HCG are optimized in the latent space to adapt the pre-trained Video DiT to compression reconstruction and learn complementary guidance conditions:
\begin{equation}
  Loss_2 = MSE(\hat{L},L)
\end{equation}
Here, $\hat{L}$ denotes the latent slices enhanced by OSDiT.

In the third stage, U2LC, HCG, and OSDiT are jointly optimized. To achieve optimal performance, this stage is conducted in the pixel domain and incorporates perceptual quality terms, including LPIPS~\cite{LPIPS} and DISTS~\cite{DISTS}:
\begin{equation}
  \begin{aligned}
    Loss_3 =& R + \lambda (\lambda_1 MSE(\hat{X},X) + \\
    &\lambda_2 LPIPS(\hat{X},X) + \lambda_3 DISTS(\hat{X},X))
  \end{aligned}
\end{equation}
where $X$ and $\hat{X}$ are the original and reconstructed videos, and $\lambda_1$, $\lambda_2$, and $\lambda_3$ balance the quality terms.

\begin{figure*}[!t]
  \centering
  \includegraphics[width=2\columnwidth]{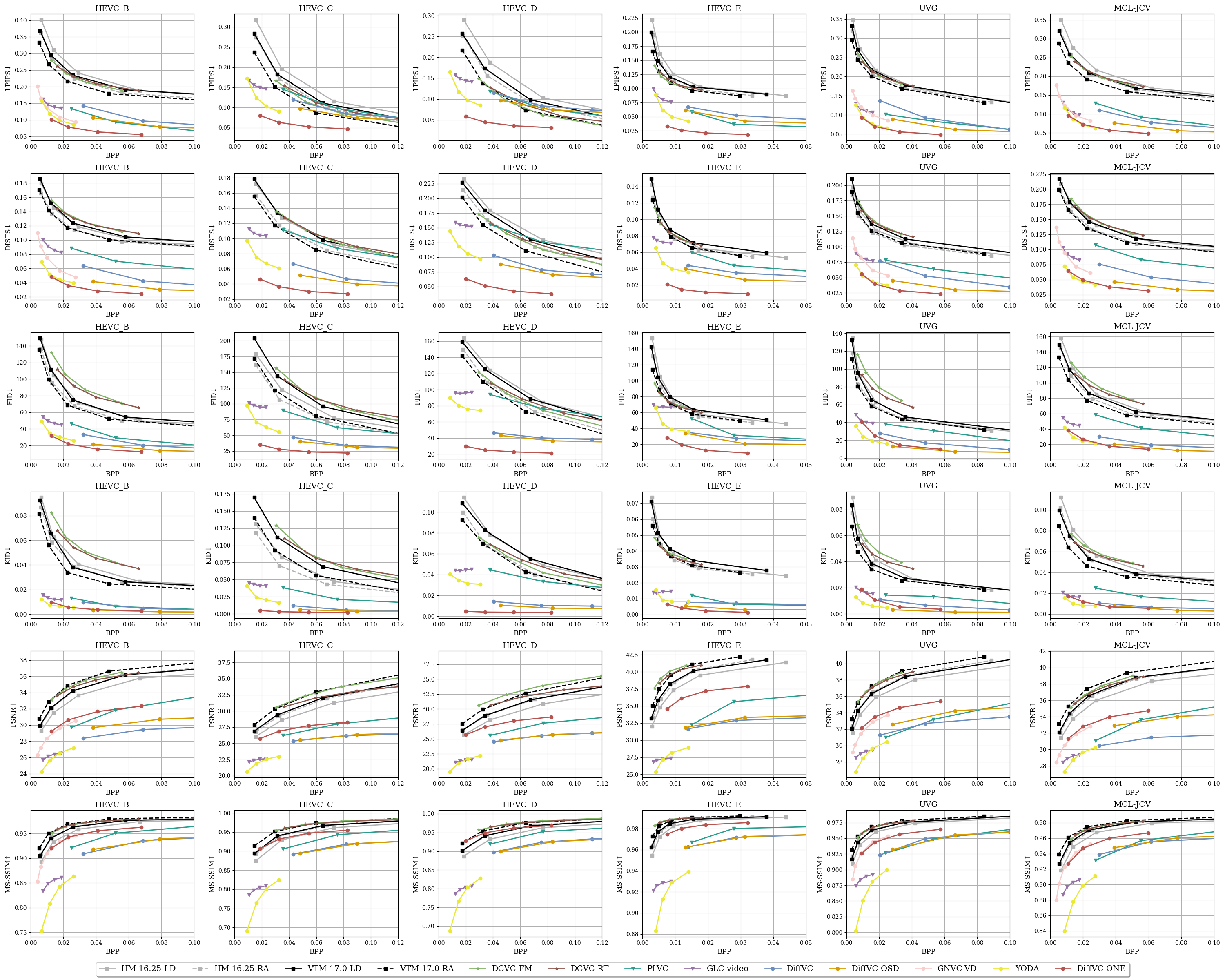}
  \caption{
    Rate-perception/distortion curves of our DiffVC-ONE and other methods on the HEVC, MCL-JCV, and UVG datasets.
  }
  \label{fig:rd_curve}
\end{figure*}

\begin{figure*}[!t]
  \centering
  \includegraphics[width=2\columnwidth]{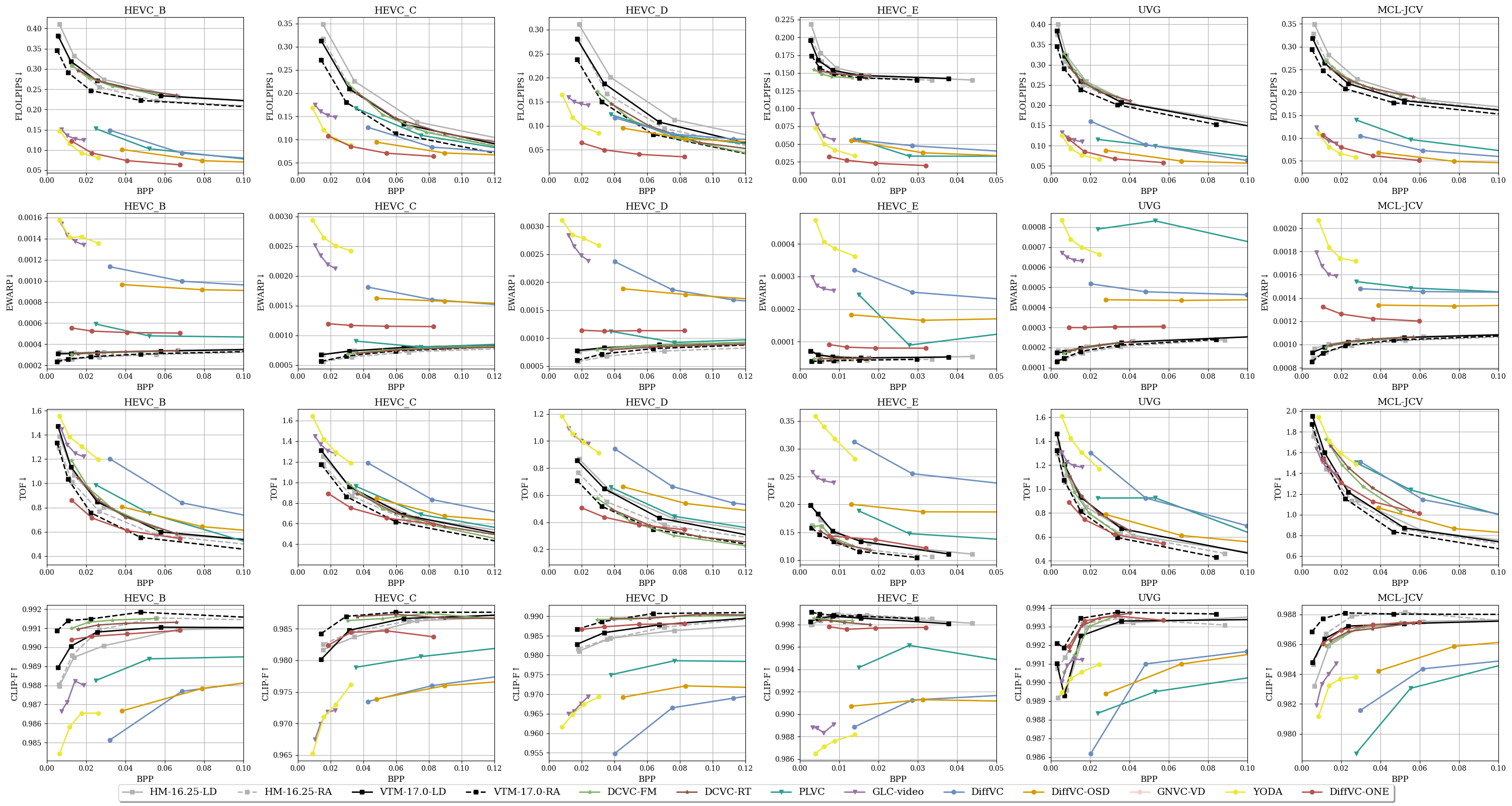}
  \caption{
    Rate-consistency curves of our DiffVC-ONE and other methods on the HEVC, MCL-JCV, and UVG datasets.
  }
  \label{fig:rc_curve}
\end{figure*}

\begin{table}[t]
  \caption{Complexity comparison in terms of computation, parameters, and encoding/decoding speed across resolutions and devices. ``-'' denotes data is unavailable.}
  \label{tab:complexity}
  \centering
  \scriptsize   
  \setlength{\tabcolsep}{4pt}
  \renewcommand{\arraystretch}{1.05}

  \begin{tabular*}{\columnwidth}{@{\extracolsep{\fill}}lcccc@{}}
  \toprule
  \multirow{2}{*}{\textbf{Method}} & \multirow{2}{*}{\shortstack{\textbf{Total}\\\textbf{Params (M)}}} & \multirow{2}{*}{\textbf{kMACs/pixel}} & \multicolumn{2}{c}{\textbf{Enc./Dec. Speed (fps)}} \\ \cmidrule(lr){4-5}
  & & & 480p on RTX 3090 & 1080p on A800 \\ \midrule
  DiffVC              & 1867.64 & 110083.01 & 0.22 / 0.23 & 0.02 / 0.02 \\
  DiffVC-OSD          & 1519.67 & 9393.59   & 2.16 / 2.78 & 0.67 / 0.87 \\
  GNVC-VD             & 2334.50 & -         & -           & 6.54 / 0.64 \\
  YODA                & 2050.27 & 10567.71  & 1.74 / 2.53 & 0.88 / 1.36 \\
  \textbf{DiffVC-ONE}  & \textbf{2302.69} & \textbf{8921.12} & \textbf{12.72 / 5.05} & \textbf{5.31 / 1.63} \\ \bottomrule
  \end{tabular*}
\end{table}

\section{Experiments}\label{sec:experiments}
\subsection{Experimental Setup}
\subsubsection{Training Settings}
We train DiffVC-ONE on 36,971 high-quality clips from OpenVid-HD~\cite{OpenVid} using Adam~\cite{Adam} with $\beta_1=0.9$ and $\beta_2=0.999$. The three stages are trained for 120K, 60K, and 108K steps with batch sizes of 24, 6, and 2, respectively. All training patches are cropped to $256\times256$.

\subsubsection{Evaluation Settings} 
We evaluate DiffVC-ONE on HEVC Classes B--E~\cite{HEVC}, UVG~\cite{UVG}, and MCL-JCV~\cite{MCL-JCV}. All sequences are converted to RGB using BT.709, and the first 96 frames are tested. We report BPP for bitrate; PSNR and MS-SSIM~\cite{MS-SSIM} for distortion; LPIPS~\cite{LPIPS}, DISTS~\cite{DISTS}, FID~\cite{FID}, and KID~\cite{KID} for perception; and Ewarp~\cite{Ewarp}, FloLPIPS~\cite{FloLPIPS}, tOF~\cite{tOF}, and CLIP-F~\cite{CLIP-F} for temporal consistency.

\subsubsection{Compared Methods} 
We compare DiffVC-ONE with representative methods, including the traditional codecs HM-16.25~\cite{HEVC} and VTM-17.0~\cite{VVC} under Low-Delay (LD) and Random-Access (RA) configurations; distortion-oriented NVC methods DCVC-FM~\cite{DCVC-FM} and DCVC-RT~\cite{DCVC-RT}; GAN-based perceptual NVC methods PLVC~\cite{PLVC} and GLC-video~\cite{GLC-video}; and diffusion-based NVC methods DiffVC~\cite{DiffVC}, DiffVC-OSD~\cite{DiffVC-OSD}, GNVC-VD~\cite{GNVC-VD} and YODA~\cite{YODA}. 

\subsubsection{Implementation Details}
We adopt Wan-2.1 VACE 1.3B~\cite{Wan} as the pre-trained Video DiT and fine-tune it using LoRA~\cite{LoRA}. For the HCG strength condition, the base, minimum, and maximum normalized noise positions are set to 0.1, 0.0, and 0.3, respectively, and mapped to the UniPC~\cite{UniPC} schedule to obtain $\sigma_{\mathrm{base}}$, $\sigma_{\min}$, and $\sigma_{\max}$. We set $\lambda_1=\lambda_2=\lambda_3=1.0$ in the loss $Loss_3$. During inference, we process GOPs of 32 frames and perform actual entropy coding.

\subsection{Main Results}
\subsubsection{Performance Comparison} 
\begin{figure*}[!t]
  \centering
  \includegraphics[width=2\columnwidth]{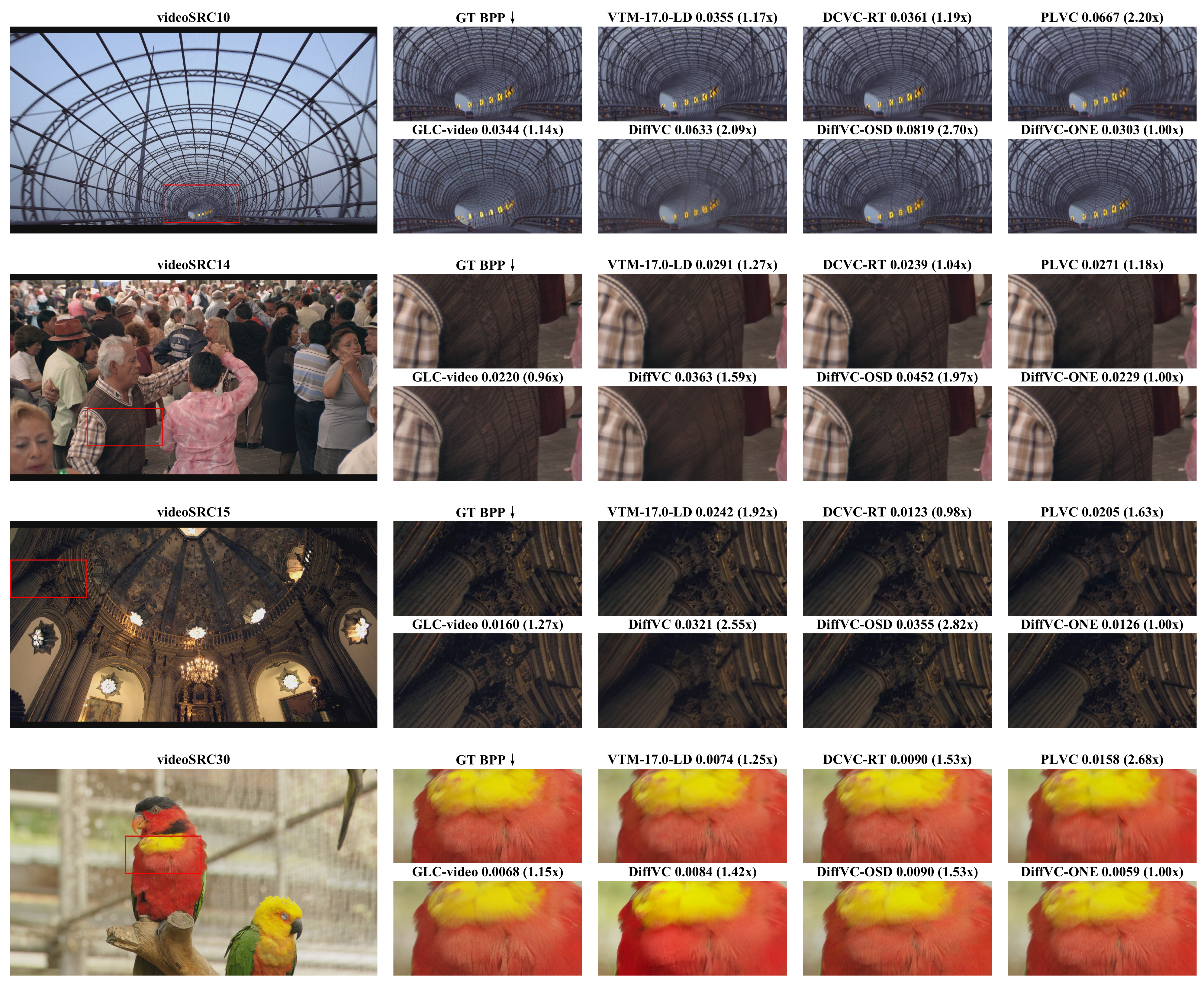}
  \caption{
    Visual comparison between the proposed DiffVC-ONE and other representative methods.
  }
  \label{fig:main_visual_result}
\end{figure*}

For quantitative comparison, Fig.~\ref{fig:rd_curve} compares the rate-perception/distortion performance of different methods across multiple datasets. DiffVC-ONE achieves the best results on nearly all perceptual metrics, with particularly strong gains in LPIPS and DISTS. It also outperforms other generative methods on distortion metrics, although it remains behind distortion-oriented codecs, consistent with the rate--distortion--perception trade-off. For qualitative comparison, Fig.~\ref{fig:main_visual_result} shows reconstructions produced by representative methods. At low bitrates, traditional and distortion-oriented methods tend to produce blurry reconstructions, while existing generative methods may introduce visually rich but unfaithful textures. In contrast, DiffVC-ONE preserves high fidelity while recovering rich perceptual details.

\subsubsection{Consistency Comparison} 
\begin{figure*}[!t]
  \centering
  \includegraphics[width=2\columnwidth]{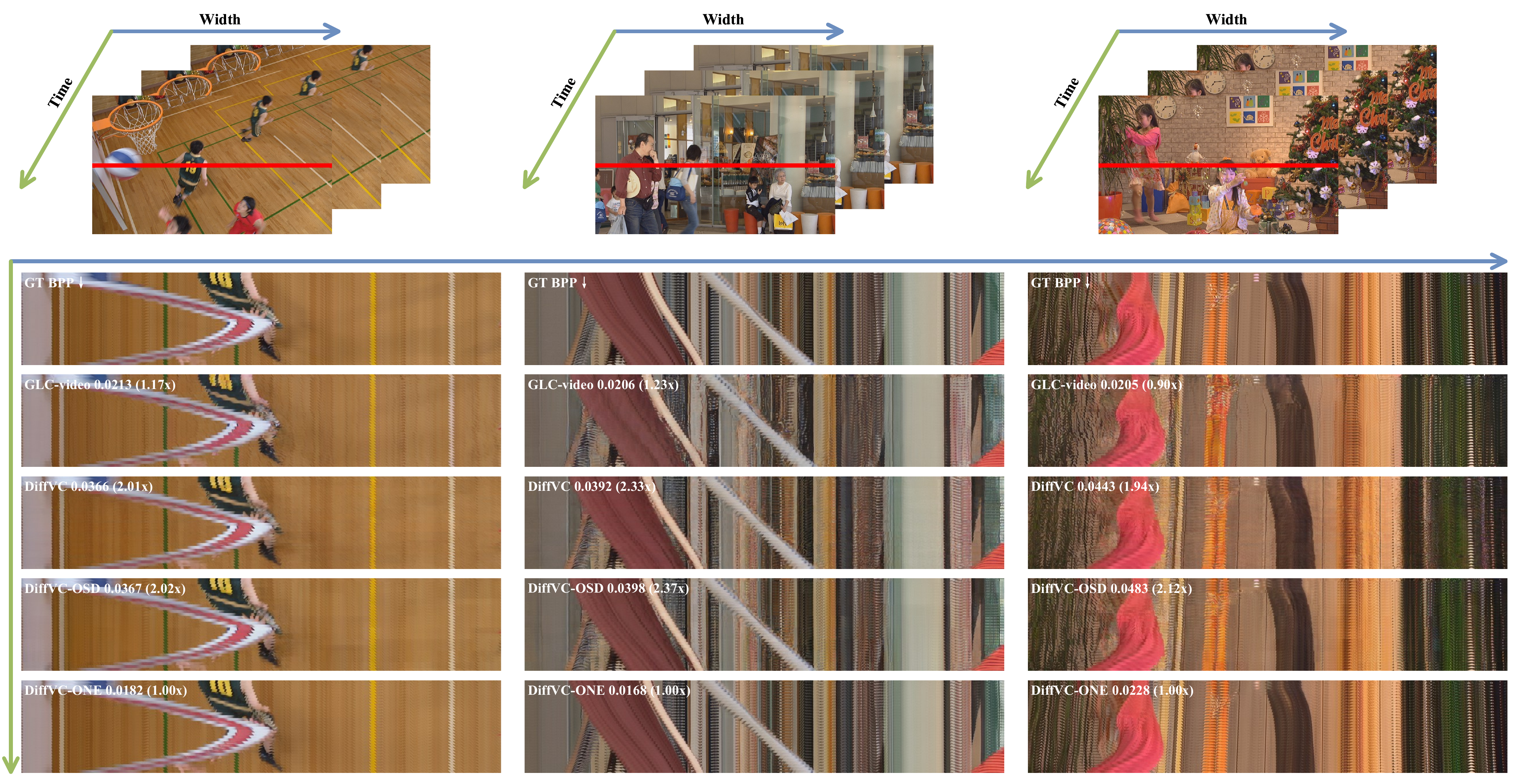}
  \caption{
    Temporal consistency comparison of generative NVC methods, visualized by stacking red-line regions across consecutive frames. From left to right: BasketballDrill, BQMall, and PartyScene from HEVC Class C. Please zoom in for details.
  }
  \label{fig:consistency_visual_result}
\end{figure*}

Fig.~\ref{fig:rc_curve} presents rate--consistency comparisons across multiple datasets. DiffVC-ONE consistently ranks among the best methods in FloLPIPS and tOF and outperforms other generative codecs in Ewarp and CLIP-F. The visual results in Fig.~\ref{fig:consistency_visual_result} further show that GLC-video, DiffVC, and DiffVC-OSD suffer from noticeable temporal flickering, whereas DiffVC-ONE achieves more stable results by one-step enhancement of multiple frames.

\begin{table}[t]
  \caption{
    Detailed complexity analysis of DiffVC-ONE. The reported Time denotes the average per-frame runtime measured on 480p video sequences using a single RTX 3090 GPU.
  }
  \label{tab:complexity_detail}
  \centering
  \scriptsize   
  \setlength{\tabcolsep}{3pt}
  \renewcommand{\arraystretch}{1.05}

  \begin{tabular*}{1.0\columnwidth}{@{\extracolsep{\fill}}lcccccc@{}}
  \toprule
  \textbf{DiffVC-ONE} & \textbf{VAE Encoder} & \textbf{U2LC} & \textbf{HCG} & \textbf{OSDiT} & \textbf{VAE Decoder} & \textbf{ALL} \\
  \midrule
  Params (M) & 53.60 & 20.87 & 0.95 & 2153.97 & 73.30 & 2302.69 \\
  kMACs/pixel & 2539.89 & 54.84 & 3.31 & 2052.33 & 4270.74 & 8921.12 \\
  Time (s/frame) & 0.0732 & 0.0050 & 0.0002 & 0.0667 & 0.1254 & 0.2706 \\ \bottomrule
  \end{tabular*}
\end{table}

\subsubsection{Complexity Analysis}
Table~\ref{tab:complexity} compares the complexity of diffusion-based NVC methods. Although DiffVC-ONE contains more parameters due to its larger pre-trained generative model, GOP-level one-step diffusion reduces the computation per pixel and improves decoding speed. Table~\ref{tab:complexity_detail} further breaks down the complexity of each component, which shows U2LC and HCG remain lightweight, while most computation is concentrated in OSDiT and the VAE.

\begin{table*}[!t]
  \caption{Ablation studies of DiffVC-ONE. All performance results are reported as BD-Rate$\downarrow$ (\%) and BD-Metric$\uparrow$ on HEVC Class C with DiffVC-ONE as the anchor. 'N/A' indicates that BD-Rate cannot be calculated due to the lack of overlap.}
  \label{tab:ablation}
  \centering
  \scriptsize  
  \resizebox{1.0\textwidth}{!}{
  \begin{tabular}{@{}lcccccccccccc@{}}
  \toprule
  \multirow{2}{*}{\textbf{Model Variants}} & \multicolumn{8}{c}{\textbf{Perception}} & \multicolumn{4}{c}{\textbf{Distortion}} \\
  \cmidrule(lr){2-9}\cmidrule(lr){10-13}
  & \multicolumn{2}{c}{\textbf{LPIPS}} & \multicolumn{2}{c}{\textbf{DISTS}} & \multicolumn{2}{c}{\textbf{FID}} & \multicolumn{2}{c}{\textbf{KID}} & \multicolumn{2}{c}{\textbf{PSNR}} & \multicolumn{2}{c}{\textbf{MS-SSIM}} \\ \midrule
  \textbf{A}: Base & \multicolumn{2}{c}{273.52 / -0.0339} & \multicolumn{2}{c}{264.77 / -0.0258} & \multicolumn{2}{c}{298.21 / -16.4871} & \multicolumn{2}{c}{N/A / -0.0072} & \multicolumn{2}{c}{256.06 / -2.1798} & \multicolumn{2}{c}{153.43 / -0.0347}  \\
  \textbf{B}: A + U2LC & \multicolumn{2}{c}{260.05 / -0.0342} & \multicolumn{2}{c}{252.20 / -0.0235} & \multicolumn{2}{c}{273.77 / -15.2428} & \multicolumn{2}{c}{N/A / -0.0061} & \multicolumn{2}{c}{247.86 / -2.2003} & \multicolumn{2}{c}{161.96 / -0.0369}  \\
  \textbf{C}: B + OSDiT & \multicolumn{2}{c}{24.66 / -0.0053} & \multicolumn{2}{c}{30.65 / -0.0037} & \multicolumn{2}{c}{33.88 / -2.7871} & \multicolumn{2}{c}{101.11 / -0.0017} & \multicolumn{2}{c}{12.19 / -0.2032} & \multicolumn{2}{c}{6.45 / -0.0021}  \\
  \textbf{D}: C + Structural Condition & \multicolumn{2}{c}{9.03 / -0.0020} & \multicolumn{2}{c}{9.73 / -0.0013} & \multicolumn{2}{c}{20.55 / -1.6972} & \multicolumn{2}{c}{61.11 / -0.0010} & \multicolumn{2}{c}{7.14 / -0.1202} & \multicolumn{2}{c}{4.12 / -0.0013}  \\
  \textbf{E}: D + Strength Condition & \multicolumn{2}{c}{-1.51 / 0.0003} & \multicolumn{2}{c}{2.13 / -0.0003} & \multicolumn{2}{c}{3.44 / -0.3084} & \multicolumn{2}{c}{20.19 / -0.0004} & \multicolumn{2}{c}{-2.43 / 0.0415} & \multicolumn{2}{c}{-1.69 / 0.0005}  \\
  \textbf{DiffVC-ONE}: E + Semantic Condition & \multicolumn{2}{c}{0.00 / 0.0000} & \multicolumn{2}{c}{0.00 / 0.0000} & \multicolumn{2}{c}{0.00 / 0.0000} & \multicolumn{2}{c}{0.00 / 0.0000} & \multicolumn{2}{c}{0.00 / 0.0000} & \multicolumn{2}{c}{0.00 / 0.0000} \\ \bottomrule
  \end{tabular}
  }
\end{table*}

\begin{table*}[t]
  \caption{
    Performance, complexity, and flexibility comparison of the Latent Compressor under different reference structures. Performance reports the BD-Rate$\downarrow$ (\%) of perception and distortion metrics on HEVC Class C, with the unidirectional variant as the anchor. Time denotes the average inference latency of the Latent Compressor for a 480p frame using a single RTX 3090 GPU. Variable Length indicates whether video inputs of varying lengths are supported.
  }
  \label{tab:latent_compressor_ablation}
  \centering
  \scriptsize   
  \setlength{\tabcolsep}{4pt}
  \renewcommand{\arraystretch}{1.05}

  \begin{tabular*}{2.0\columnwidth}{@{\extracolsep{\fill}}cccccccccccc@{}}
  \toprule
  \multirow{2}{*}{\textbf{Reference Structure}} & \multicolumn{7}{c}{\textbf{Performance}} & \multicolumn{4}{c}{\textbf{Complexity \& Flexibility}} \\
  \cmidrule(lr){2-8} \cmidrule(lr){9-12}
  & \textbf{LPIPS} & \textbf{DISTS} & \textbf{FID} & \textbf{KID} & \textbf{PSNR} & \textbf{MS-SSIM} & \textbf{Average} & \textbf{Params (M)} & \textbf{kMACs/pixel} & \textbf{Time (s/frame)} & \textbf{Variable Length} \\ \midrule
  Unidirectional & 0.00 & 0.00 & 0.00 & 0.00 & 0.00 & 0.00 & 0.00 & 20.87 & 54.84 & 0.0050 & True \\
  Joint & 25.27 & 29.27 & 25.72 & 22.75 & 29.77 & 24.59 & 26.23 & 25.11 & 63.24 & 0.0047 & False \\ 
  Bidirectional & 10.59 & 12.21 & 9.57 & 9.63 & 11.55 & 11.42 & 10.83 & 30.40 & 62.43 & 0.0058 & False \\ \bottomrule
  \end{tabular*}
\end{table*}

\subsection{Ablation Studies}\label{sec:ablation}
To validate the design of DiffVC-ONE, we conduct comprehensive incremental ablation studies, as summarized in Table~\ref{tab:ablation}. Detailed analyses are provided below.

\subsubsection{Unified Unidirectional Latent Compressor}\label{sec:ablation_u2lc}
Model A, which combines a separate latent compressor with a multi-step Video DiT, serves as the baseline. Model B replaces the latent compressor with U2LC. The results show that the unified latent compressor achieves better overall performance than the separate design while reducing unnecessary parameter redundancy.
Furthermore, Table~\ref{tab:latent_compressor_ablation} compares the Latent Compressor under three reference structures. The joint variant employs a costly 3D architecture and models an entire GOP with a single network, resulting in higher complexity and clearly inferior performance to the unidirectional one. The bidirectional variant also performs worse, with an average BD-Rate of 10.83\%. Without explicit motion modeling, large displacements between hierarchical references weaken temporal context prediction under limited complexity. In contrast, the unidirectional variant offers a better balance between compression performance and complexity. Its recursive reference structure also generalizes more reliably from short training clips to longer test sequences, whereas joint and bidirectional variants are more sensitive to GOP-length mismatches. We therefore adopt unidirectional variant as the default mode.

\subsubsection{Video DiT-based One-Step Diffusion Enhancer}
Model C replaces the multi-step Video DiT in Model B with OSDiT. The results show that one-step diffusion enhancement yields a substantial performance gain. Unlike multi-step sampling, which creates a long backpropagation path and makes end-to-end training prohibitively expensive, OSDiT enables joint pixel-domain optimization and substantially improves compression performance.

\subsubsection{Hybrid Condition Generator}
Comparing Models D and C shows that the structural condition improves both perceptual and distortion metrics by providing spatial layouts, object boundaries, and motion cues. Adding the strength condition in Model E yields a further average gain of 15.26\% by adapting the enhancement intensity to the video content and compression level. Finally, introducing the semantic condition produces the complete DiffVC-ONE and further improves overall performance, including a 20.19\% gain in KID, demonstrating the benefit of high-level semantics for distribution alignment. Together, the three complementary conditions provide effective guidance for one-step diffusion and achieve the best compression performance.

\begin{table}[t]
  \caption{
    Impact of LoRA rank on BD-Rate$\downarrow$ (\%) for perception and distortion metrics on HEVC Class C, using rank 32 as the anchor.
  }
  \label{tab:lora_rank_ablation}
  \centering
  \scriptsize   
  \setlength{\tabcolsep}{4pt}
  \renewcommand{\arraystretch}{1.05}

  \begin{tabular*}{1.0\columnwidth}{@{\extracolsep{\fill}}cccccccc@{}}
  \toprule
  \multirow{2}{*}{\textbf{LoRA Rank}} & \multicolumn{4}{c}{\textbf{Perception}} & \multicolumn{2}{c}{\textbf{Distortion}} & \multirow{2}{*}{\textbf{Average}} \\
  \cmidrule(lr){2-5} \cmidrule(lr){6-7} 
  & \textbf{LPIPS} & \textbf{DISTS} & \textbf{FID} & \textbf{KID} & \textbf{PSNR} & \textbf{MS-SSIM} & \\ \midrule
  16 & -0.82 & 5.50 & 5.66 & 11.27 & 0.44 & 1.21 & 3.88 \\
  32 & 0.00 & 0.00 & 0.00 & 0.00 & 0.00 & 0.00 & 0.00 \\
  64 & -3.34 & 0.30 & 1.90 & 7.36 & -3.95 & -2.48 & -0.04 \\ \bottomrule
  \end{tabular*}
\end{table}

\subsubsection{LoRA Rank}
We further ablate the effect of the LoRA rank in OSDiT on compression performance, as summarized in Table~\ref{tab:lora_rank_ablation}. Increasing the rank from 16 to 32 improves the average performance by 3.88\%, whereas a further increase to 64 yields only a marginal 0.04\% gain. We therefore use a LoRA rank of 32 as the default setting.

\section{Conclusion}\label{sec:conclusion}
This paper presents DiffVC-ONE, a generative video compression framework built on a one-step Video Diffusion Transformer. DiffVC-ONE jointly performs temporal compression, condition modeling, and generative enhancement in a compact video latent space. U2LC efficiently compresses latent slices using a shared coding architecture, while OSDiT exploits the spatiotemporal generative prior of a pre-trained Video DiT to enhance an entire GOP in a single step, substantially reducing sampling cost while maintaining cross-frame consistency. In addition, the HCG provides complementary structural, strength, and semantic guidance, enabling more accurate recovery of information lost during compression. Extensive experiments across multiple standard benchmarks demonstrate that DiffVC-ONE achieves state-of-the-art perceptual quality together with strong temporal consistency.

\section{Acknowledgement}
The numerical calculations in this paper have been done on the supercomputing system in the Supercomputing	Center of Wuhan University.

\bibliographystyle{IEEEtran}
\bibliography{reference}

\vfill

\end{document}